\documentclass[submission,copyright,creativecommons]{eptcs}
\providecommand{\event}{ICLP 2023} 

\usepackage{iftex}

\usepackage{amsmath}
\usepackage{graphicx}
\usepackage{multirow}

\usepackage{iftex}
\usepackage{algorithm} 
\usepackage{algpseudocode} 

\usepackage{subfig}
\usepackage{graphicx}
\usepackage{adjustbox}

\usepackage{xcolor}
\usepackage{soul}
\usepackage[normalem]{ulem}

\usepackage{bbm}
\usepackage{amsmath}
\usepackage{amsfonts}
\DeclareMathOperator*{\argmax}{arg\,max}

\usepackage{graphicx}
\usepackage{siunitx}
\usepackage{multirow}

\usepackage{booktabs}

\usepackage{tikz}

\usepackage{hyperref}

\ifpdf
  \usepackage{underscore}         
  \usepackage[T1]{fontenc}        
\else
  \usepackage{breakurl}           
\fi

\title{Deep Inductive Logic Programming meets Reinforcement Learning}
\author{Andreas Bueff
\institute{University of Edinburgh}
\and
Vaishak Belle
\institute{University of Edinburgh}
}
\def\titlerunning{Deep Inductive Logic Programming meets Reinforcement Learning}
\def\authorrunning{Andreas Bueff, Vaishak Belle}
\begin{document}
\maketitle

\begin{abstract}
One approach to explaining the hierarchical levels of understanding within a machine learning model is the symbolic method of inductive logic programming (ILP), which is data efficient and capable of learning first-order logic rules that can entail data behaviour. A differentiable extension to ILP, so-called differentiable Neural Logic (dNL) networks, are able to learn Boolean functions as their neural architecture includes symbolic reasoning. We propose an application of dNL in the field of Relational Reinforcement Learning (RRL) to address dynamic continuous environments. This represents an extension of previous work in applying dNL-based ILP in RRL settings, as our proposed model updates the architecture to enable it to solve problems in continuous RL environments. 
The goal of this research is to improve upon current ILP methods for use in RRL by incorporating non-linear continuous predicates, allowing RRL agents to reason and make decisions in dynamic and continuous environments.

\end{abstract}


\section{Introduction}
One approach to explaining the hierarchical levels of understanding within a machine learning model is the symbolic method of inductive logic programming (ILP) \cite{
MUGGLETON1994629}, which is data efficient and capable of learning first-order logic rules that can entail data behaviour.
Recent contributions to the field have expanded the ILP framework to allow for end-to-end learning, resulting in hybrid models that can be classified as neuro-symbolic \cite{evansnoisydata, payanidNL,yang2019learn, dongneurallogicmachine}. 
The recent developments in neuro-symbolic ILP have expanded the potential applications of these models to a wider range of learning challenges, including reinforcement learning \cite{payani2020incorporating}. 
The ILP-based neuro-symbolic model implemented in this proposal is a differentiable extension to ILP, so-called differentiable Neural Logic (dNL) networks \cite{payanidNL}. dNL networks are able to learn Boolean functions as their neural architecture includes symbolic reasoning. The primary neural layers in the dNL model contain weighted neurons associated with conjunctions as well as weighted activation neurons associated with disjunctions, providing a means of logical reasoning on the input as well as a means of optimisation via gradient descent.

Relational RL (RRL) is concerned with learning policies for decision-making tasks in complex, discrete relational environments \cite{Driessens2010}. In RRL, the environment is characterised by a set of entities and relationships between them, and the agent's actions can affect the state of these entities and relationships. 
Due to its focus on relational problems, RRL has leveraged the progress made in ILP but is not limited to one method. Recent efforts have enabled RRL to effectively tackle more complex RL problems by transitioning from purely symbolic reasoning to a more neuro-symbolic approach that leverages neural systems. Earlier research in RRL primarily focused on planning \cite{Driessens2010}, emphasising model-based learning. Later research, however, focused on model-free methods that combined neuro-symbolic models and RRL \cite{zambaldi2018deep} or on the derivation of interpretable FOL policies \cite{jiang2019neural, payani2020incorporating}. Despite these advancements, there is limited research in applying RRL to more complex learning challenges, such as continuous state dynamics. This gap in knowledge motivates the proposed research.


The work by Payani et al. extended RRL to learn FOL policies using a dNL agent \cite{payani2020incorporating}. Taking the concepts of dNL, Payani et al. combined their dNL-ILP framework with RRL \cite{payani2020incorporating}. The author tested on the block world gaming environment and took advantage of the declarative bias with provided background knowledge. The authors expanded RRL to handle complex scene interpretations and used their dNL-ILP differentiable inductive engine to give RRL an end-to-end learning framework, so-called dNL-RRL, where the authors focused on policy gradients in order to improve interpretability and expert constraints to improve the speed of convergence. 

The following paper is an extension that incorporates both a continuous and non-linear interpretation of \cite{payanidNL} and the dNL-RRL agent found in \cite{payani2020incorporating} resulting in a dNL-ILP based agent that can both learn in dynamic continuous environments typically seen in RL control problems and extract FOL rules which define the agent policy. Various RL algorithms were explored in evaluating our dNL-ILP agent and ultimately Soft Actor-Critic was found to perform the best when solving problems on continuous state spaces. As the contribution bridges the application of dNL-ILP agents in discrete state spaces with that of continuous, we evaluated the model on a dynamic RL environment where the optimal policy captured rules such as those seen with classical mechanics in physics. In our evaluation, the model was applied to two control problems including the Cart Pole problem and Lunar Lander problem \cite{openAIgym}. From our initial results, we were able to obtain agent policies which incorporate continuous predicate functions as well as non-linear continuous predicate functions. We also add that our proposed agent, while primarily focused on solving classical RL problems, is referred to as an RRL agent due to its use of relational language in the derived FOL rules and the ability to incorporate background knowledge through the use of non-linear predicate functions.


\section{Background}
\subsection{Reinforcement Learning}

Reinforcement learning (RL) aims to find an optimal action sequence for an agent to maximize its expected future reward. This is typically done by modelling the environment as a Markov Decision Process (MDP), which is defined as a tuple $\mathcal{M} = \langle \mathcal{S}, \mathcal{A}, \mathcal{R}, \mathcal{T}, \mathcal{E} \rangle$. $\mathcal{S}$ is the set of states, $\mathcal{A}$ is the set of actions that can be taken, $\mathcal{R}(s_t,a_t, s_{t+1})$ is the reward function that takes the current state $s_t$, current action $a_t$, and returns the reward from transitioning to the state $s_{t+1}$, $\mathcal{T}(s_t, a_t, s_{t+1})$ is the transition probability function which represents the probability of transitioning to state $s_{t+1}$ from state $s_t$ given action $a_t$ was taken, and $\mathcal{E} \subset \mathcal{S}$ is the set of terminal states \cite{sutton1998introduction,leikescalableagent}. A discount factor $\gamma \in [0,1]$ determines the importance of receiving a reward in the current state versus the future state, where $R_t = \sum_{k=0}^{\infty}\gamma^k r(a_{t+k},s_{t+k})$ is the total accumulated return from the time step. The value function $V^{\pi}(s) = \mathbb{E}_{\pi}[R_t]$ of a policy is the measure of the expected sum of discounted rewards. The goal is to find the optimal policy $\pi^* = \argmax_{\pi} \mathbb{E}[\sum_{t=0}^{\infty} \gamma^t r(a_t,s_t) | s_0 = s]$, which maps a history of observations to the next action\cite{sutton1998introduction,rodrigueztowards,leikescalableagent,roderick2017deep,pmlr-v80-icarte18a,A2C}. In practice, at each time step, the agent selects an action based on its current state and the policy, and receives a reward based on the transition to the next state. In our investigation of RL methods, we found that Soft Actor-Critic (SAC) is the best approach for our dNL based agent. SAC combines the strengths of both value-based and policy-based methods by alternating between updates to the policy and updates to the value function and Q-function. The entropy regularisation term is added to the policy update step to encourage exploration, and the algorithm uses a temperature parameter, $\alpha$, to control the trade-off between exploration and exploitation. As we evaluate our agent on environments consisting of discrete actions, we use a discrete action variant of SAC \cite{SACdiscrete}. The main difference is in the policy update step, where the objective becomes $\pi^* = \argmax_{\pi}\sum_{t = 0}^T \mathbb{E}_{s_t \sim D} [  \mathbb{E}_{a_t \sim \pi}[r(s_t,a_t) + \alpha \mathbb{H}(\pi(\cdot|s_t))]] $.
 Where $\mathbb{H}(\pi(a|s))$ is the entropy of the policy, $\gamma$ is the discount factor, $D$ is the replay buffer, and $\alpha$ is the temperature parameter. In the discrete setting, $\pi$ maps states to a vector of probabilities with $|A|$ elements. For the actor cost function, we use the following equation $J_{\pi}(\phi) = \mathbb{E}_{s_t \sim D} [  \mathbb{E}_{a_t \sim \pi}[\alpha \log(\pi_{\phi}(a_t|s_t))-Q_{\theta}(s_t, a_t)]]$.
Where $\pi_\phi$ is the policy parameterized by $\phi$ and $Q_\theta$ is the Q-function parameterized by $\theta$. Discrete SAC policy maximises the probability of discrete actions as opposed to the continuous SAC policy which optimises two parameters of a Gaussian distribution.

\subsection{Inductive Logic Programming (ILP)}

    
    

Inductive logic programming (ILP) is a method of symbolic computing which automatically constructs logic programs given a background knowledge base (KB) \cite{MUGGLETON1994629}. An ILP problem is represented as a tuple $(\mathcal{B}, \mathcal{P}, \mathcal{N})$ of ground atoms, with $\mathcal{B}$ being the background assumptions, $\mathcal{P}$ being a set of positive instances which help define the target predicate to be learned, and $\mathcal{N}$ being the set of negative instances of the target predicate. The goal of ILP is to construct a logic program that explains all provided positive sets and rejects the negative ones. Given an ILP problem $(\mathcal{B}, \mathcal{P}, \mathcal{N})$, the goal is to identify a set of hypotheses (clauses) $\mathcal{H}$ such that $\mathcal{B} \wedge \mathcal{H} \models \gamma$ for all $\gamma \in \mathcal{P}$ and $\mathcal{B} \wedge \mathcal{H} \not\models \gamma$ for all $\gamma \in \mathcal{N}$, where $\models$ denotes logical entailment. In other words, the conjunction of the background knowledge and hypothesis should entail all positive instances and not entail any negative instances.

\subsection{Differentiable Neural Logic (dNL)}
 The core component of the dNL network is their use of differentiable neural logic layers to learn Boolean functions \cite{payanidNL}. The dNL architecture uses membership weights and conjunctive and disjunctive layers to learn a target predicate or Boolean function. Learning a target predicate $p$ requires the construction of Boolean function $\mathcal{F}_p$ which passes in a Boolean vector $\textbf{x}$ of size $N$ with elements $x^{(i)}$, into a neural conjunction function $f_{conj}$ (see equation \ref{fconj}) which is defined by a conjunction Boolean function $F_{conj}$ (see equation \ref{boolconj}). A predicate defined by Boolean function in this matter is extracted by parsing the architecture for membership weights ($w^{(i)}$) above a given threshold, where membership weights are converted to Boolean weights via a sigmoid $m^{(i)} = \sigma(c w^{(i)})$ with constant $c \geq 1$. Membership weights are paired with continuous lower and upper bound predicate functions (see equation \ref{conPred1}) which are eventually interpreted as atoms in the body of the predicate being learned. These same Boolean predicate functions are used to transform non-Boolean data into a Boolean format for the logic layers.
\begin{subequations}
             \begin{equation}\label{fconj}
            f_{conj}(\textbf{x}) = \prod_{i=1}^N F_{conj}(x^{(i)}, m^{(i)})
        \end{equation}
        
        \begin{equation}\label{boolconj}
            F_{conj}(x^{(i)}, m^{(i)}) = \overline{\overline{x^{(i)}} m^{(i)}} = 1 - m^{(i)}(1 - x^{(i)})
        \end{equation}
\end{subequations}

 Following induction by the conjunctive layer, outputs are fed into a neural  disjunction function $f_{disj}$ (see equation \ref{fdisj}) which is constructed from disjunctive Boolean functions $F_{disj}$ (see equation \ref{booldisj}). The disjunctive layer provides multiple definitions for our target predicate if necessary.
\begin{subequations}
                    \begin{equation}\label{fdisj}
            f_{disj}(\textbf{x}) = 1 - \prod_{i =1}^N(1 - F_{disj}(x^{(i)}, m^{(i)}))
        \end{equation}
        
        \begin{equation}\label{booldisj}
          F_{disj}(x^{(i)}, m^{(i)}) = x^{(i)} m^{(i)}  
        \end{equation}
\end{subequations}

\begin{figure}
    \centering
    \begin{tikzpicture}
  
  \node[draw,minimum width=2cm,minimum height=0.5cm] (xstar) at (3.5,0.8) {$x^* = 0.8$};
  
  \node[draw,minimum width=1.5cm,minimum height=0.5cm,fill=pink!20] (f1) at (0.75,0) {$\mathcal{F}_{lt_x^1}(x,0.5)$};
  \node[draw,minimum width=1.5cm,minimum height=0.5cm,fill=blue!20] (f2) at (2.75,0) {$\mathcal{F}_{lt_x^2}(x,1.0)$};
  \node[draw,minimum width=1.5cm,minimum height=0.5cm,fill=blue!20] (f3) at (4.77,0) {$\mathcal{F}_{gt_x^1}(x,0.5)$};
  \node[draw,minimum width=1.5cm,minimum height=0.5cm,fill=pink!20] (f4) at (6.83,0) {$\mathcal{F}_{gt_x^2}(x,1.0)$};
  
  \draw[->] (xstar.south) -- (f1.north);
  \draw[->] (xstar.south) -- (f2.north);
  \draw[->] (xstar.south) -- (f3.north);
  \draw[->] (xstar.south) -- (f4.north);
  
\end{tikzpicture}
    \caption{{Given $k=2$ and feature instance $x^*$ with value $0.8$, would result in two distinct activations of bounded Boolean predicates. Stating} \textit{$x^*$ is less than $1.0$ and $x^*$ is greater than $0.5$.}}
    \label{fig:dNLpredlabel}
\end{figure}
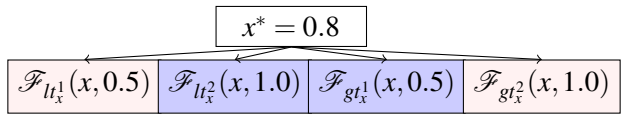

As mentioned, a target predicate function $\mathcal{F}p$ is defined by the cascading architecture of the dNL network, and within the network bounded continuous Boolean predicates associated with a continuous feature are used to define our target predicate. Boolean predicate functions are used to handle continuous input by partitioning input into a series of lower and upper-bound predicates, referred to as continuous predicates as they capture the interval bounds of continuous features. Similarly, Boolean predicates mapped to discrete features are referred to as discrete predicates. These bounded continuous predicates return either true or false when a continuous value meets the condition. A continuous input $x$ is associated with $k$ pairs of upper and lower boundary predicates, each pair corresponding to a bounded range $ (l_{xi} < x < u_{xi})$, where $i \in {1,2,\cdots,k }$. A Boolean upper boundary predicate $gt_x^i(x,l{xi})$, states whether \textit{``x is greater than $l_{xi}$"} is true, and a Boolean lower boundary predicate $lt_x^i(x,u_{xi})$ states whether \textit{``x is less than $u_{xi}$"} is true (see equation \ref{conPred1}). The lower and upper boundary values $l_{xi}$ and $u_{xi}$ are treated as trainable weights which can be optimised during training. To clarify the need for both lower and upper predicate bounds, the regions are complete and disjoint, meaning that each input value belongs to exactly one region. Although most of the upper and lower bounds may coincide, having both upper and lower predicates allows for greater flexibility in defining the regions and ensures that no input values fall outside the defined regions. A diagram illustrating this partitioning of the input domain can be found in Figure \ref{fig:dNLpredlabel}.

            \begin{equation}\label{conPred1}
            \mathcal{F}_{gt^i_{x}} = \sigma(c(x-u_{xi})), \quad \mathcal{F}_{lt^i_{x}} = \sigma(-c(x-l_{xi}))
        \end{equation}

\section{Methodology}

Developing a neuro-symbolic RL agent that incorporates continuous predicate functions posed several non-trivial challenges. Firstly, the work of Payani et al. \cite{payanidNL, payani2020incorporating}, which our model extends, did not explore non-linear predicate functions, so we had to develop new functions to capture more complex relationships in the data (see equation \ref{trnpred}). Additionally, while Payani et al. did introduce continuous predicates in a supervised setting, they only tested dNL-ILP on two datasets (Wine and Sonar datasets \cite{payanidNL, Dua:2019}), whereas we made the integration of continuous predicates a focus of our work. \footnote{Moreover, from an engineering standpoint, the dNL code was originally written in TensorFlow 1, which had to be converted to TensorFlow 2 for use in the RL environment.} {Another challenge was the fact that the original dNL-RRL model explicitly dealt with discrete predicates, so we had to adapt the model to incorporate continuous predicates.} Finally, while Payani et al. used the REINFORCE RL algorithm \cite{reinforceSutton}, we had to test out various other RL algorithms, including SAC \cite{SAC}, to find the most effective implementation for our model.\footnote{Extensive coding was required to convert the SAC algorithm, which typically has a discrete action space, to TensorFlow 2, as all available examples online were coded in PyTorch.}

In the dNL-RL interpretation, the Boolean target predicate function $\mathcal{F}_p$ is referred to as a \textit{discrete action predicate function}. The actions of the agent are the target predicates and so must remain discrete. For example, an agent that can take an action $turnLeft()$, has the corresponding \textit{discrete action predicate function} $\mathcal{F}_{turnLeft}$. Each \textit{discrete action predicate function} is defined by an input matrix $\mathbf{I}$, which is composed of continuous lower and upper bound predicate functions as well as discrete Boolean predicate functions, all with associated weights. The set of the continuous state features $cnt_p$ and the set of the discrete state features $dsc_p$ for the associated RL environment are given Boolean interpretations after being passed through the input matrix, as seen in equation \ref{inputmatrix}.{ The row dimensions of $\mathbf{I}$ are defined by the batch size $\mathbf{b}$ and the column dimensions are defined by $N_e = (2\times k \times |cnt_p|) + (2 \times |dsc_p|)$ (assuming discrete features are Boolean)}. { The state feature ranges used to define the Boolean predicate functions} are determined by equal-width binning, and we define the batch size as $\mathbf{b}$. The $\mathcal{F}_{{e}}$ function takes in a one-hot encoded Boolean discrete input.

\begin{equation}\label{inputmatrix}
\mathbf{I} = \begin{bmatrix}
\mathcal{F}_{gt^1_{e_1}},\mathcal{F}_{lt^1_{e_1}}  & \cdots & \mathcal{F}_{gt^k_{e_1}},\mathcal{F}_{lt^k_{e_1}}  & \cdots & \mathcal{F}_{gt^k_{e_{|cnt_p|}}},\mathcal{F}_{lt^k_{e_{|cnt_p|}}}\\

\vdots& \vdots & \vdots & \vdots & \vdots
\end{bmatrix} \bigoplus
\begin{bmatrix}
\mathcal{F}_{{e_1}},\mathcal{F}_{\overline{e_1}} & \cdots & \mathcal{F}_{{e_{|dsc_p|}}},\mathcal{F}_{\overline{e_{|dsc_p|}}} \\
\vdots & \vdots & \vdots \

\end{bmatrix}
\end{equation}

In defining the \textit{discrete action predicate function} $\mathcal{F}_p$ for an action $p$, the function takes as input the input matrix $\mathbf{I}$ as well as the disjunction layer $F_{disj}$ and the conjunction layer $F_{conj}$ where $N_e$ is used to define the number of layers in the conjunction layer and $N_p$ is used to define the number of layers in the disjunction layer, seen in equation \ref{booltargetpredfunc}. As an agent learns rules for multiple actions, we define the set of discrete action predicates as the \textit{predicate action policy} $\pi_{\mathcal{F}}$ which is the set $\{ \mathcal{F}_{p^1}, \mathcal{F}_{p^2}, \cdots, \mathcal{F}_{p^n}  \}$ for all actions in the action space $p \in \mathcal{A}$ for the RL environment. 

{To prevent the gradient from becoming excessively small during training, initial weights $\mathbf{W}$ are initialised using a negative mean random Gaussian distribution, which ensures that they are close to zero. To keep membership weights $\textbf{m}$ between 0 and 1, a constant $c \geq 1$ is applied, followed by a sigmoid function. In equations} \ref{congj} and \ref{disj}, {we apply the conjunction function to the input matrix using the corresponding membership weights. Note that $\textbf{W}^{conj}$ is a matrix with dimensions $N_p$ and $N_e$ while $\textbf{W}^{disj}$ is a vector of size $N_p$. The disjunction function takes the output of the conjunction function as input as we cascade our neural architecture.}

\begin{equation}\label{booltargetpredfunc}
\qquad \quad \mathcal{F}_p|_{\mathbf{I},N_e,N_p} = F_{disj}(N_p, F_{conj}(N_e, \mathbf{I} ))      
\end{equation}

\begin{subequations}
\begin{equation}\label{congj}
  \quad {F}_{conj} = \sum_{i=1}^{N_e} [1 - m^{conj}_i(1- \mathbf{I}_i) ] \quad 
\text{where} \quad  \mathbf{m}^{conj} = \sigma(c\mathbf{W}^{conj}), \ \mathbf{W}^{conj}_{N_p, N_e} \sim \mathcal{N}   
\end{equation}

\begin{equation}\label{disj}
 \quad {F}_{disj} = 1 - \sum_{i=1}^{N_p} [1 - m_i^{disj}{F}_{conj}^i ] \quad 
\text{where} \quad  \mathbf{m}^{disj} = \sigma(c\mathbf{W}^{disj}), \ \mathbf{W}^{disj}_{1, N_p} \sim \mathcal{N}
\end{equation}
\end{subequations}

We incorporate non-linear transformations on some of the state features. As in the case of various control problems in RL, the calculation of state transitions is based on non-linear transformations from the current state. We define the following $k$ upper-boundary and lower-boundary functions as non-linear transformation predicates in equation \ref{trnpred}.
\begin{equation}\label{trnpred}
    \mathcal{F}_{gt^i_{f(x)}} = \sigma(c(f(x)-u_{xi})), \: \mathcal{F}_{lt^i_{f(x)}} = \sigma(-c(f(x)-l_{xi})) \: \text{where} \: f(x) \in \{sqr(x), sin(x), \cdots\}
\end{equation}


Our dNL based agents can be differentiated by the type of prior information used. The baseline dNL agent uses only continuous and discrete Boolean predicates and is referred to as a dNL RL continuous (dNLRLc) agent. It learns the continuous state features without using any prior information from a knowledge base $\mathsf{KB}_{T}$ which contains transformation functions mapped to specific continuous state features. However, when prior information such as non-linear equations or transformations between features is included, the agent becomes a dNL RL non-linear continuous (dNLRLnlc) agent. This involves adding non-linear continuous Boolean predicates to the input matrix as new predicate functions in $\pi_{\mathcal{F}}$. In the experiment section, we specify which non-linear transformations were used for the dNLRLnlc agent in each RL environment.

In Algorithm \ref{dNLagent} we provide a pseudo-code implementation for a dNL-RL agent. The agent takes in a state  $\mathbf{s}_t$,  at time step $t$ and returns an action $\mathbf{a}_t$ for that time step. The algorithm iterates through each feature of the state, $s_t^{(i)}$, and separates the elements into two cases: discrete and continuous. If the element is discrete, it adds the mapping of the discrete predicate $dsc_p[i]$ to the state element $s_t^{(i)}$ in a processed state set $\mathbb{S}_t$. If the element is continuous, it checks if a transformation function for the feature exists in the knowledge base $\mathsf{KB}_{T}[i]$. If it does, it adds the mapping of the continuous predicate $cnt_p[i]$ to the function applied to the state feature $s_t^{(i)}$ in the state set $\mathbb{S}_t$. If the knowledge base does not exist, it adds the mapping of the continuous predicate $cnt_p[i]$ to the state feature $s_t^{(i)}$ instead. After iterating through all state features, the policy is returned by applying the dNLRL policy function to the set $\mathbb{S}_t$, and then sampling an action $\textbf{a}_t$ from the \textit{predicate action policy}.

	\begin{algorithm} \scriptsize
	\caption{dNL-RL agent} 
 \scriptsize
	\begin{algorithmic}[1]
	\State \textbf{Input}: $\mathbf{s}_t$-state at time step $t$
	\State \textbf{Output}: $\mathbf{a}_t$-action at time step $t$
	
	\For{$s_t^{(i)} \in \mathbf{s}_t$}
                \If{ $s_t^{(i)}$ is discrete }
                    \State $\mathbb{S}_t
                    \cup \{ dsc_p[i] \mapsto  s_t^{(i)}\}$
                \EndIf

                \If{ $s_t^{(i)}$ is continuous}
                    \If{$\mathsf{KB}_{T}[i]$ exists}
                        \State $\mathbb{S}_t \cup \{ cnt_p[i] \mapsto  \mathsf{KB}_{T}[i](s_t^{(i)})\}$
                    \Else
                        \State $\mathbb{S}_t \cup \{ cnt_p[i] \mapsto  s_t^{(i)}\}$
                    \EndIf
                \EndIf

	\EndFor
        \State $\mathbb{\pi}_\mathcal{F} \leftarrow \mathrm{dNLRL}(\mathbb{S}_t)$ 
        \State $\mathbf{a}_t \sim \mathbb{\pi}_\mathcal{F}(\mathcal{F}_{a_t}|\mathbb{S}_t)$

	\end{algorithmic} \label{dNLagent}
\end{algorithm}

To derive the predicate action policy $\mathbb{\pi}_\mathcal{F}$, we perform Boolean reasoning on the set of processed state features with associated predicates as seen in Algorithm \ref{dNLRLpolicy}.  Here we take as input a set of discrete predicates $dsc_p$, a set of continuous predicates $cnt_p$, a set of target action predicates $\mathbb{P}$, and a set of processed state feature predicates $\mathbb{S}_t$. We loop through each target action predicate in the set $\mathbb{P}$ and for each predicate, it creates two empty sets used for constructing the input matrix, one for the continuous predicates $\mathbf{I}_c$ and one for the discrete predicates $\mathbf{I}_d$. The dNLRL policy then loops through each continuous predicate in $cnt_p$ and for each predicate, it creates a disjunction of predicates, which are added to the continuous input matrix set $\mathbf{I}_c$. Then it does the same for each discrete predicate in $dsc_p$ but instead, the discrete predicates are added to the discrete input matrix set $\mathbf{I}_d$. We take the union of these two sets to get the final input matrix $\mathbf{I}$. Finally, it creates an evaluated target action predicate $\mathcal{X}_p$ by reasoning on target predicate function $\mathcal{F}_p$ as defined in equation \ref{booltargetpredfunc}. This target action predicate $\mathcal{X}_p$ is then added to the predicate action policy $\mathbb{\pi}_\mathcal{F}$.

\begin{algorithm}
	\caption{Single Step Forward Chain Model/ dNLRL policy} 
  \scriptsize
	\begin{algorithmic}[1]
	\State \textbf{Input}: $dsc_p$-set of discrete predicates, $cnt_p$-set of continuous predicates, $\mathbb{P}$-set of target action predicates, $\mathbb{S}_t$-set of state feature predicates at time step $t$
	\State \textbf{Output}: $\mathbb{\pi}_\mathcal{F}$-predicate action policy

	    \For {$p \in \mathbb{P}$}

	        \State $\mathbf{I}_c$=[]
                \State $\mathbf{I}_d$=[]
	        \For {$e \in cnt_p$}

	                \State $\mathbf{I}_c \cup \bigvee^{k}_{i=1} \Big( \mathcal{F}_{gt^i_{e}}(\mathbb{S}_t[e]) \vee \mathcal{F}_{lt^i_{e}}(\mathbb{S}_t[e]) \Big)$

	       \EndFor

                \For {$e \in dsc_p$}

	                \State $\mathbf{I}_d \cup  \Big( \mathcal{F}_{e}(\mathbb{S}_t[e]) \vee \mathcal{F}_{\overline{e}}(\mathbb{S}_t[e]) \Big)$

	       \EndFor
              \State $\mathbf{I} \leftarrow \mathbf{I}_d \cup \mathbf{I}_c$

	       \State $\mathcal{X}_p \leftarrow \mathcal{F}_p|_{\mathbf{I},N_e,N_p}$

	       \State $\mathbb{\pi}_\mathcal{F} \cup (\mathcal{X}_p)$
	    \EndFor

	\end{algorithmic}  \label{dNLRLpolicy}
\end{algorithm}

\section{Experiments and Results}
\subsection{Environments and Tasks}

The following RL environments are continuous control problems developed by Open AI Gym \cite{openAIgym}. The Cart Pole problem is a benchmark for evaluating RL algorithms \cite{Lytkin2005ReinforcementLB}. The goal of the problem is to balance a pole in the upright position by selecting between two discrete actions (move left, move right). The state space is 4-dimensional, all continuous. The features are the following $x:$ Cart Position, $x':$ Cart Velocity, $\theta:$ Pole Angle, and $\theta':$ Pole Angular Velocity. In the Lunar Lander environment, an agent learns a policy for throttling the side and main engines to control the descent of a lander such that it lands on a landing pad. The discrete actions include activating the right thruster engine, activating the left thruster engine, activating the main engine, and doing nothing. The state space is 8-dimensional, 6 inputs are continuous and 2 are discrete. The continuous state inputs include including $x:$ Lander Position on the X-axis, $y:$ Lander Position on the Y-axis, $v_x:$ Horizontal velocity, $v_y:$ Vertical velocity, $\theta:$ Angular orientation in space, and $v_{\theta}:$ Angular velocity. The remaining two features are Boolean, Left: which indicates if the left leg is touching the ground and Right: which indicates if the right leg is in contact with the ground.

\subsubsection{Baselines}
During our investigation into the development of our continuous logic-based algorithm, we sought first to see if the dNLRLc agent could in fact learn interpretable policies on continuous environments. As stated this led to evaluation of the dNLRLc architectures on various RL algorithms. This includes the policy gradient algorithm REINFORCE \cite{reinforceSutton}, the model-free algorithm Deep Q-Network (DQN) \cite{DQN}, the on-policy method of Advantage Actor-Critic (A2C) \cite{A2C}, and the Soft Actor-Critic (SAC) \cite{SAC}. In the case of SAC, we use a variant designed to handle discrete actions \cite{SACdiscrete}. In cases where we inject prior knowledge in the form of non-linear functions added to the transformation knowledge base, we designate the model as (SAC-NL).

\subsection{Algorithm Performance Comparison}
We present algorithmic comparisons of the various RL algorithms combined with the dNLRLc agent where evaluation is carried out on the Cart Pole problem. For the evaluation, only SAC is used to test the dNLRLnlc  agent, as it was the best-performing algorithm with the dNLRLc agent. In the Lunar Lander environment, we evaluate a dNLNLc agent and dNLRLnlc agent using SAC entirely. 
Each algorithmic framework used the same binning scheme for the continuous features, that being  `equal-width binning' where bins for each feature were predefined for the learning environment and kept constant between the various RL algorithms.  


\subsubsection{Task: Cart Pole Problem}
\begin{figure}[ht]
    \centering
    \includegraphics[width=10cm]{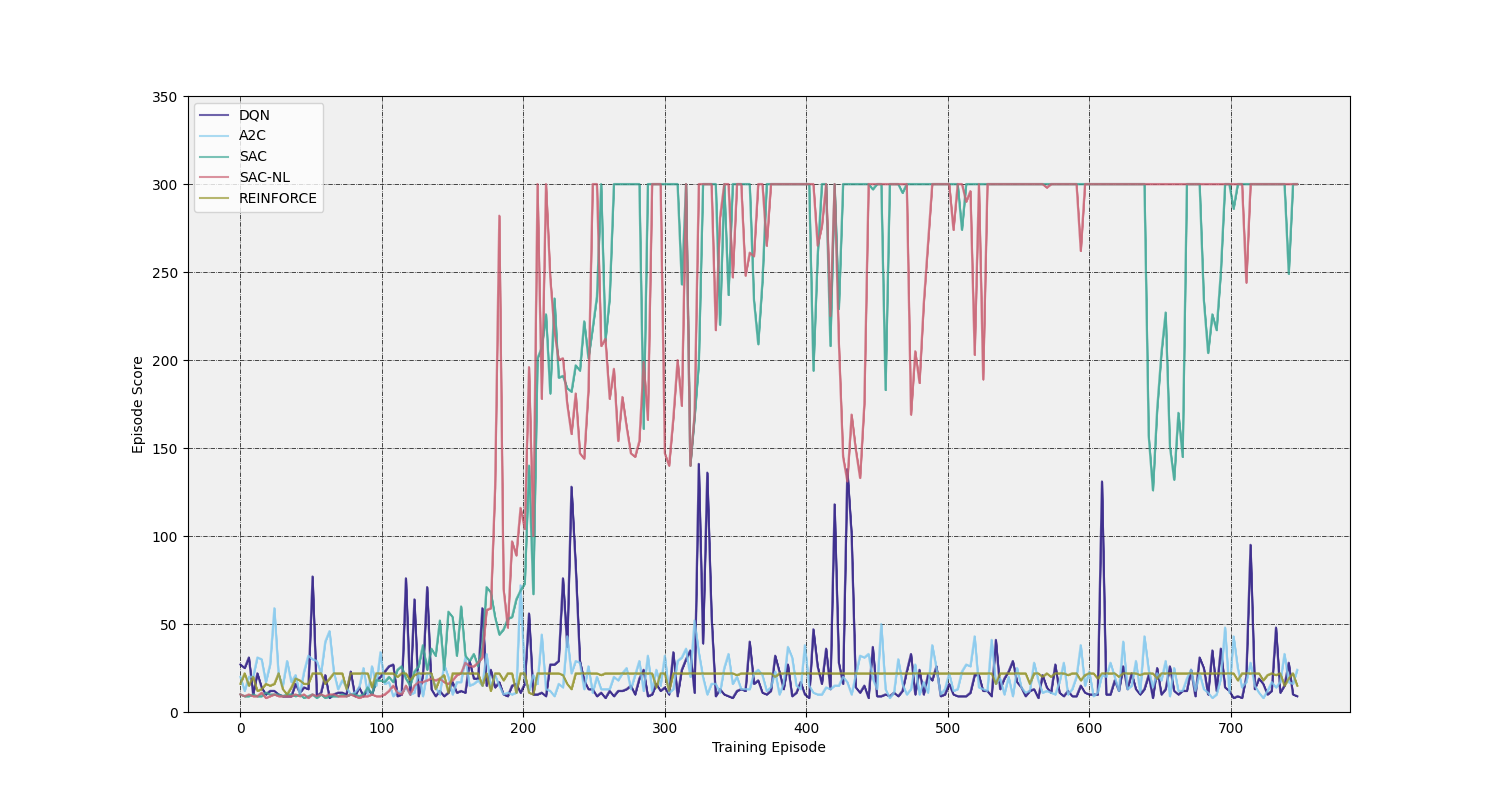}
    \caption{\scriptsize Episode scores for RL algorithms relying on dNLRLc and dNLRLnlc agents.}
    \label{fig:cartpole1}
\end{figure}

\begin{figure}
\centering
\begin{minipage}{.5\textwidth}
  \centering
  \includegraphics[width=1.2\linewidth]{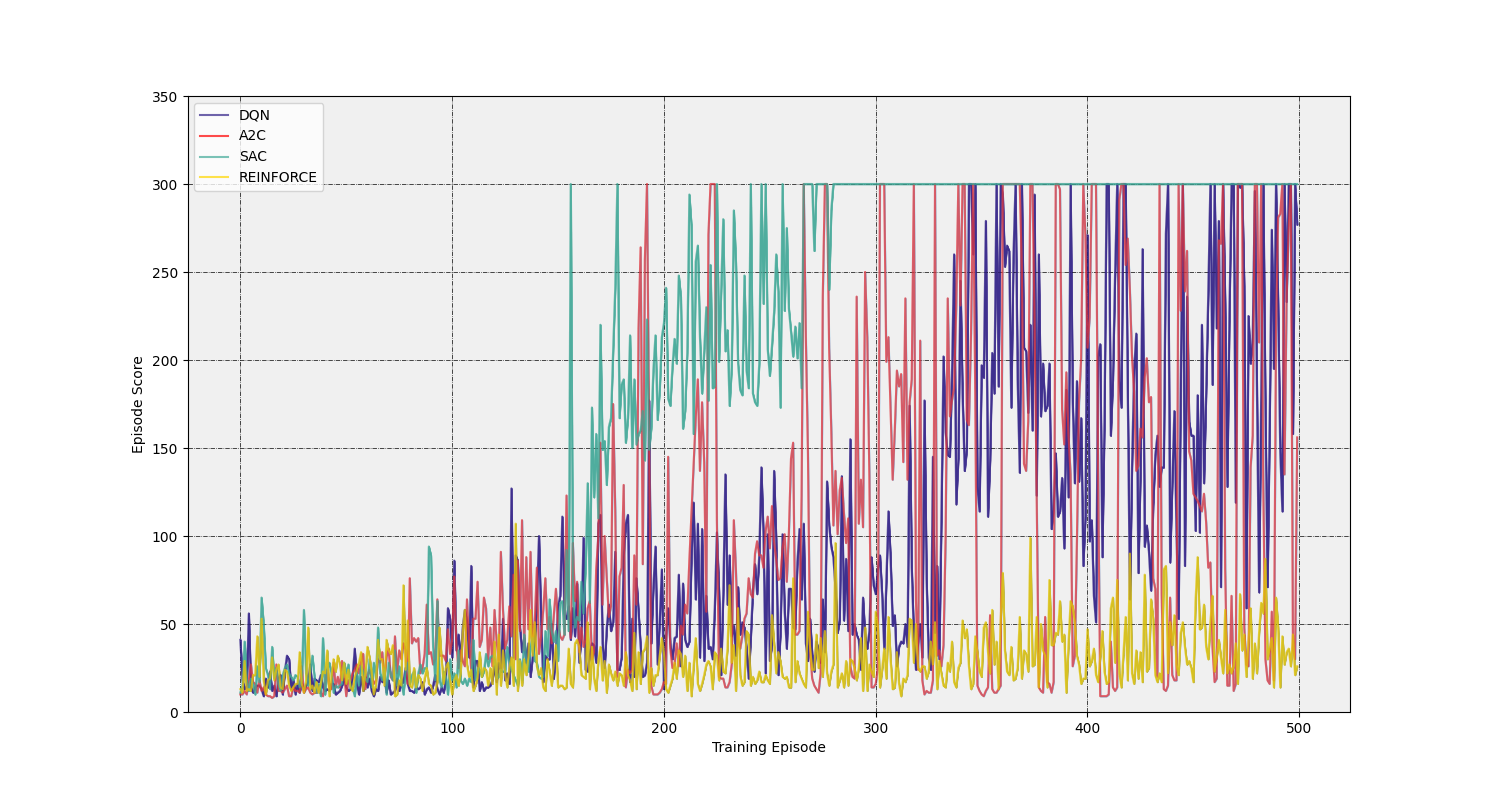}
  \captionof{figure}{ \scriptsize Episode scores for RL algorithms relying on neural network based agents.}
  \label{fig:cartpole1bb}
\end{minipage}%
\begin{minipage}{.5\textwidth}
  \centering
  \includegraphics[width=.6\linewidth]{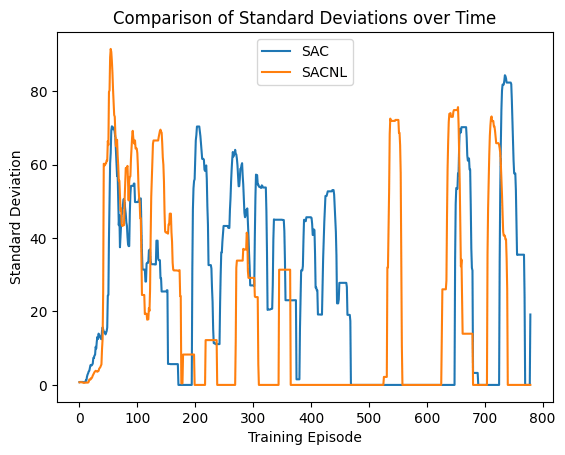}
  \captionof{figure}{\scriptsize Moving (window size 20) standard deviation for SAC relying on dNLRLc and dNLRLnlc agents.}
  \label{fig:cartpoleSTD}
\end{minipage}
\end{figure}

For the Cart Pole problem, each algorithmic framework used the same binning scheme for `equal-width binning' with the number of bins set to $[4,4,4,4]$ for each feature respectively.
In Figure \ref{fig:cartpole1}, we can observe the results comparing the performances of each dNLRLc agent and paired RL architecture in the Cart Pole Problem domain. As can be observed, the best performing model is that of the SAC and SAC-NL. Performance of dNL based agents is noticeably poor in the cases of the REINFORCE algorithm where rewards cap at $22.0$. DQN and A2C perform better but rewards plateau far below what is to be expected. The non-linear addition for the dNLRLnlc agent is a simple transformation of the Pole Angle $\theta$, as the calculation of the next state $s_{t+1}$ within the code for the Cart Pole problem is dependent on the $sine$ transformation of $\theta$. For SAC-NL, we apply a $sine$ transformation and so policies can contain atoms designated $PoleAngleSine$ with corresponding inequalities. The performance of the  dNLRLnlc agents is not significantly better than the dNLRLc agent when using SAC. Both agent variants are able to achieve the maximum reward of $300.0$ consistently with periodic fluctuations. 
{In figure} \ref{fig:cartpoleSTD}{, we can observe the moving standard deviation for the dNLRLc and dNLRLnlc when using SAC. Although the fluctuations in the standard deviations are not in sync, the variation between the two agents is not significant enough to conclude that one agent is more stable.}

Figure \ref{fig:cartpole1bb}{ presents a comparison of the performance of the various RL algorithms using standard neural networks in solving the Cart Pole problem. The figure shows that, while SAC is able to solve the problem and stabilise, the neural network agents learn faster and are more stable than dNLRLc agents. Both agents using A2C and DQN exhibit inconsistent performance with rewards fluctuating greatly and both the neural networks and dNLRLc agents perform poorly when using REINFORCE. These findings suggest that while dNLRLc models may offer greater interpretability, they may come at the cost of decreased performance in some RL tasks. Given that, it is also evident that SAC is the primary contributor to solving the control problem as even standard neural network based agents failed otherwise. We also note that the black-box neural network-based SAC agent outperforms the interpretable dNL-RL SAC agent in terms of speed, achieving an average convergence time of 2168.31 seconds, compared to the average convergence time of 10807.94 seconds for the dNL-RL agent, highlighting a trade-off between interpretability and learning efficiency, which could be addressed in future research by optimising the algorithm for faster convergence.}



The main advantage of using dNL agents is that returned policies are interpretable. These policies provide predicate logic interpretations for the individual discrete actions, in the case of the Cart Pole problem, we have rules for when to move \textit{left()} and when to move \textit{right()}. The policy in our case is defined as \textit{predicate action policy}, that is individual actions are treated as target predicates where the clausal body states what inequalities should be true for a state feature in order for the action to be considered by the agent. We also clarify that the clausal body is defined by a conjunction of atoms where the associated membership weight is placed before the atom in brackets, extracted from the conjunction neuron. If the membership weight is above $0.95$, i.e. the model is confident it should be in the definition, then the weight is not listed next to the atom. Similarly, for the disjunction neurons, the value in brackets before the rule represents the membership weights. 

\begin{table}[ht] \centering
\begin{tabular}{c l}
\hline
\scriptsize {\textbf{CartPole}}                                                       & \scriptsize {Policy rules for dNLRLc}                                                                                                                                                                                                                                                                                                                                                                                                                                     \\ \hline \small
\textbf{\scriptsize \begin{tabular}[c]{@{}c@{}} mean reward: \\ $290.3 \pm 32.3 $\end{tabular}} & \small \begin{tabular}[c]{@{}l@{}}

\tiny \textbf{left()}  \\
         
       \tiny $:- ([0.56]CartPos<2.83\wedge [0.64]CartVeloc>-1.19\wedge CartVeloc>0.18$ \\
       \tiny \qquad $\wedge PoleAngle>-0.62\wedge PoleAngleVeloc>-0.24 )$\\
       \tiny $:- ([0.81]CartPos<2.82\wedge PoleAngle>-0.06\wedge PoleAngleVeloc>0.08 )$\\

\tiny \textbf{right()}  \\

        \tiny  $:- (CartVeloc>-1.19\wedge PoleAngle<-0.03\wedge PoleAngleVeloc<0.34\wedge [0.56]PoleAngleVeloc<2.48 )$\\
        \tiny  $:- (CartPos<0.16\wedge [0.52]CartVeloc<2.11\wedge [0.50]PoleAngle>-0.56$\\
        \tiny \qquad$\wedge PoleAngle<0.11\wedge PoleAngleVeloc<0.25 )$

\end{tabular}                                                                                                                                                                                                                                                                                            \\ \hline
\end{tabular}
\caption{\scriptsize The FOL policy rules for the dNLRLc agent, trained using SAC. Extracted continuous inequality predicates define rules for each action.}\label{Tab:CartSAC-dNL-Crules}
\end{table}

In table \ref{Tab:CartSAC-dNL-Crules}, the predicate action policy for the dNLRLc agent trained using SAC is provided. In the body of the discrete action predicate \textit{left()}, it is observed that two definitions are returned wherein which we see both associated disjunction neuron membership weights pass the threshold. The three state features present in the second definition are Cart Position ($CartPos<2.82$), Pole Angle ($PoleAngle>-0.06$), and Pole Angular Velocity ($PoleAngleVeloc>0.08$). Similarly, the body for the discrete action predicate \textit{right()} contains two predicate definitions. In both cases, the agent is confident in either definition as indicated by the absence of a membership weight for the corresponding disjunction neuron.

\begin{table}[ht] \centering

\begin{tabular}{c l}
\hline
\scriptsize {\textbf{CartPole}}                                                       & \scriptsize {Policy rules for dNLRLnlc}                                                                                                                                     \\ \hline \small
\textbf{\scriptsize \begin{tabular}[c]{@{}c@{}}mean reward: \\ $294.7 \pm 25.8 $\end{tabular}} & \small \begin{tabular}[c]{@{}l@{}}

\tiny \textbf{left()}  \\
         
\tiny        $:- ([0.60]CartPos<2.57\wedge PoleAngleSine>0.00\wedge PoleAngleVeloc>-0.38 )$\\

\tiny \textbf{right()}  \\

 \tiny        $:- (PoleAngleSine<0.04\wedge PoleAngleVeloc<0.00 )$\\
 \tiny        $:- (CartPos<0.74\wedge [0.55]CartVeloc>-1.64\wedge CartVeloc<-0.11$ \\ 
 \tiny $\qquad \wedge PoleAngleSine<0.65\wedge[0.66]PoleAngleVeloc>-2.04\wedge PoleAngleVeloc<0.28 )
$\\

\end{tabular}                                                                                                                                                                                                                                                                                            \\ \hline
\end{tabular} 
\caption{\scriptsize The FOL policy rules for the dNLRLnlc agent, trained using SAC. Extracted continuous inequality predicates define rules for each action with the inclusion of non-linear continuous inequality predicates.}\label{Tab:CartSAC-dNL-NLCrules}
\end{table}

In table \ref{Tab:CartSAC-dNL-NLCrules}, the predicate action policy for the dNLRLnlc agent trained using SAC is provided. The definitions of the predicate action policy are comparable to those given by the dNLRLnlc agent. Note that the discrete action predicate \textit{left()} is defined by a single predicate rule. For both discrete action predicates, we see the presence of the non-linear continuous predicate in the definitions, that is $(PoleAngleSine >0.00)$ of the discrete action predicate $left()$ and $(PoleAngleSine < 0.00)$ for the first definition of $right()$ and $(PoleAngleSine < 0.65 \wedge PoleAngleSine < 0.28)$ for the second definition. In figure \ref{Tab:CartSAC-dNL-Crules} and figure \ref{Tab:CartSAC-dNL-NLCrules} the mean reward is also provided with the standard deviation. The mean reward in this case refers to the mean of the last 100 episodes. 



\subsubsection{Task: Lunar Lander}

At present, the Lunar Lander stands as a more challenging environment for the dNL-RL agents. As the previous control problem environment has demonstrated that SAC is the best performing algorithm, only SAC was tested on the Lunar Lander environments. In setting up the experiments, it was found that both the discretisation scheme and the initial high and low values for the state features had a significant impact. In the context of the Lunar Lander, a reward of $200$ indicates the lander successfully landed on the platform. The current binning scheme being deployed is $[3,3,3,3,3,3,-,-]$ for each feature, where ($-$) indicates no binning performed as it is a discrete/Boolean feature. In order to evaluate the consistency and performance of the dNLRLc and dNLRLnlc agents, multiple trials were conducted using different random seeds. We ran each experiment 5 times and the resulting policies were analysed. While only one representative policy is presented for each agent, any variations or discrepancies across the trials are discussed. The non-linear change was again a $sine$ transformation on the state feature angular orientation in space $\theta$ associated with the transformation predicate \textit{AngleSine}.

\begin{figure}[ht]
    \centering
    \includegraphics[width=10cm]{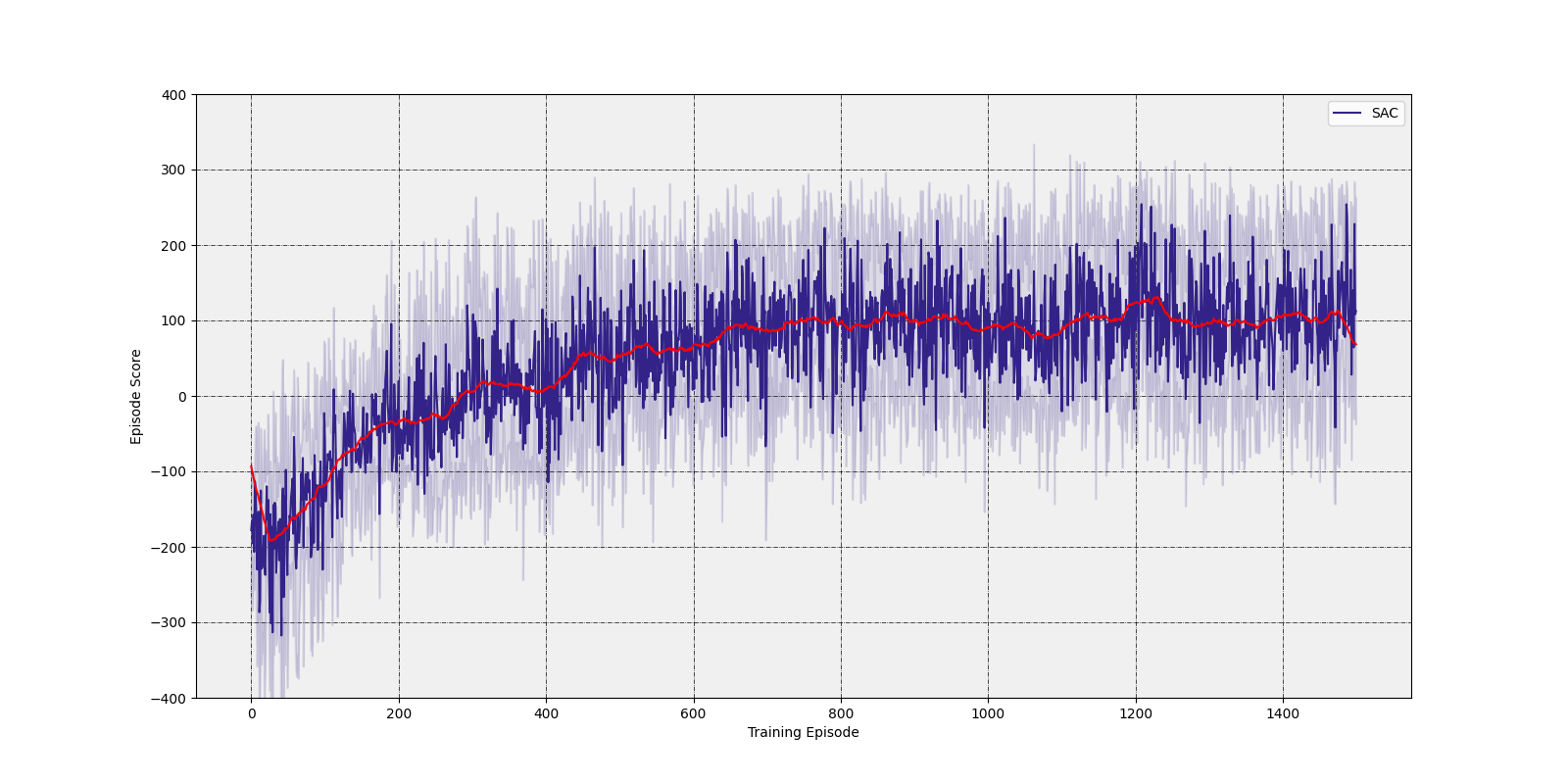}
    \caption{\scriptsize Episode scores for SAC algorithms relying on dNLRLc. Dark purple corresponds to the mean reward. Light purple corresponds to the standard deviation and the red line is the moving average for 50 episodes.}
    \label{fig:lunarlanderCseed}
\end{figure}

In Figure \ref{fig:lunarlanderCseed}, we can observe the results comparing the performance of the dNLRLc agent where dark purple corresponds to the mean reward. The results indicate the agent does in fact learn, however, it is challenging for the dNL agent to successfully land the lander consistently. The episode rewards show the agent can land the lander but the mean rewards skew below a true success of $200$. We note the mean reward across all trials fluctuates around rewards of a hundred with significant deviations from the mean. In some initial states, the learned policy does result in successful landings, often over 200. The rules in the predicate action policy for the Lunar Lander environment are given in table \ref{Tab:LunarSAC-dNL-Crules}. As the Lunar Lander has four actions, we find that all disjunction layers pass the parameter threshold and only in a few cases do we have conjunction weights for specific atoms. For example for the rule $fireLeft$, the first rule contains a weighted atom $[0.65]CoordX > -1.52$. For the action predicate, \textit{fireRight()} the agent learned four rules, where only the second rules contain listed conjunction layer weights for individual atoms $[0.60]CoordX>-1.53$ and $[0.94]AngularVeloc>-2.58$. We observe as well, that while the bounded atom $CoordX$ appears in the definitions for each action, each is associated with a conjunction weight. Indicating the agent was less confident about the inclusion of the $CoordX$ continuous predicate function.

\begin{table}[ht]\centering
\begin{tabular}{c l}
\hline
\scriptsize {\textbf{LunarLander}}                                                       & \scriptsize {Policy rules for dNLRLc}                                                                                                                               \\ \hline \small
\textbf{\scriptsize \begin{tabular}[c]{@{}c@{}}mean reward: \\ $162.1 \pm 110.9 $  \end{tabular}} & \small \begin{tabular}[c]{@{}l@{}}

\tiny \textbf{doNothing() }  \\

\tiny          $:- (CoordY<1.28\wedge LinearVelocX<-0.14 )$\\
 \tiny        $:- ([0.80]CoordX>-1.53\wedge [0.76]CoordX<1.70\wedge LinearVelocY<-0.12$\\
 \tiny \quad$\wedge[0.66]Angle>-1.60 )$\\
 \tiny         $:- ([0.82]CoordX<1.70\wedge LinearVelocY<-0.04\wedge RightLegContactFalse )$\\

\tiny \textbf{fireLeft())}  \\
\tiny          $:- ([0.65]CoordX>-1.52\wedge LinearVelocX<-0.14\wedge [0.73]Angle>-0.09 )$\\
\tiny          $:- ([0.72]CoordX<1.70\wedge [0.68]CoordY>-0.99\wedge LinearVelocX<1.93\wedge Angle>-0.09 )$\\
         
\tiny \textbf{fireMain() }  \\
 \tiny         $:- (LinearVelocX>0.20\wedge Angle<0.17\wedge AngularVeloc<0.10 )$\\
\tiny          $:- ([0.58]CoordY>-0.91\wedge LinearVelocY>-0.11 )$\\
\tiny          $:- ([0.75]CoordX<1.70\wedge LinearVelocX<-0.11\wedge Angle>0.01\wedge AngularVeloc>-0.14 )$\\

\tiny \textbf{fireRight() }  \\
\tiny          $:- (LinearVelocX>-0.63\wedge Angle>-1.60\wedge Angle<0.17\wedge AngularVeloc<-0.02 )$\\
 \tiny         $:- ([0.60]CoordX>-1.53\wedge Angle<0.17\wedge [0.94]AngularVeloc>-2.58$\\
 \tiny \quad$\wedge LeftLegContactTrue )$\\
\tiny          $:- (CoordY<1.28\wedge LinearVelocX>-0.05\wedge Angle<0.32\wedge AngularVeloc>-1.98$\\
 \tiny \quad$\wedge RightLegContactFalse )$\\
 \tiny         $:- (LinearVelocX>-0.05\wedge Angle>-1.60\wedge Angle<0.17 )$\\

\end{tabular}                                                                                                                                                                                                                                                                                            \\ \hline
\end{tabular} 
\caption{\scriptsize The FOL policy rules for the dNLRLc agent, trained using SAC. Extracted continuous bounded predicates define rules for each action.}\label{Tab:LunarSAC-dNL-Crules}
\end{table}

\begin{figure}[ht]
    \centering
    \includegraphics[width=10cm]{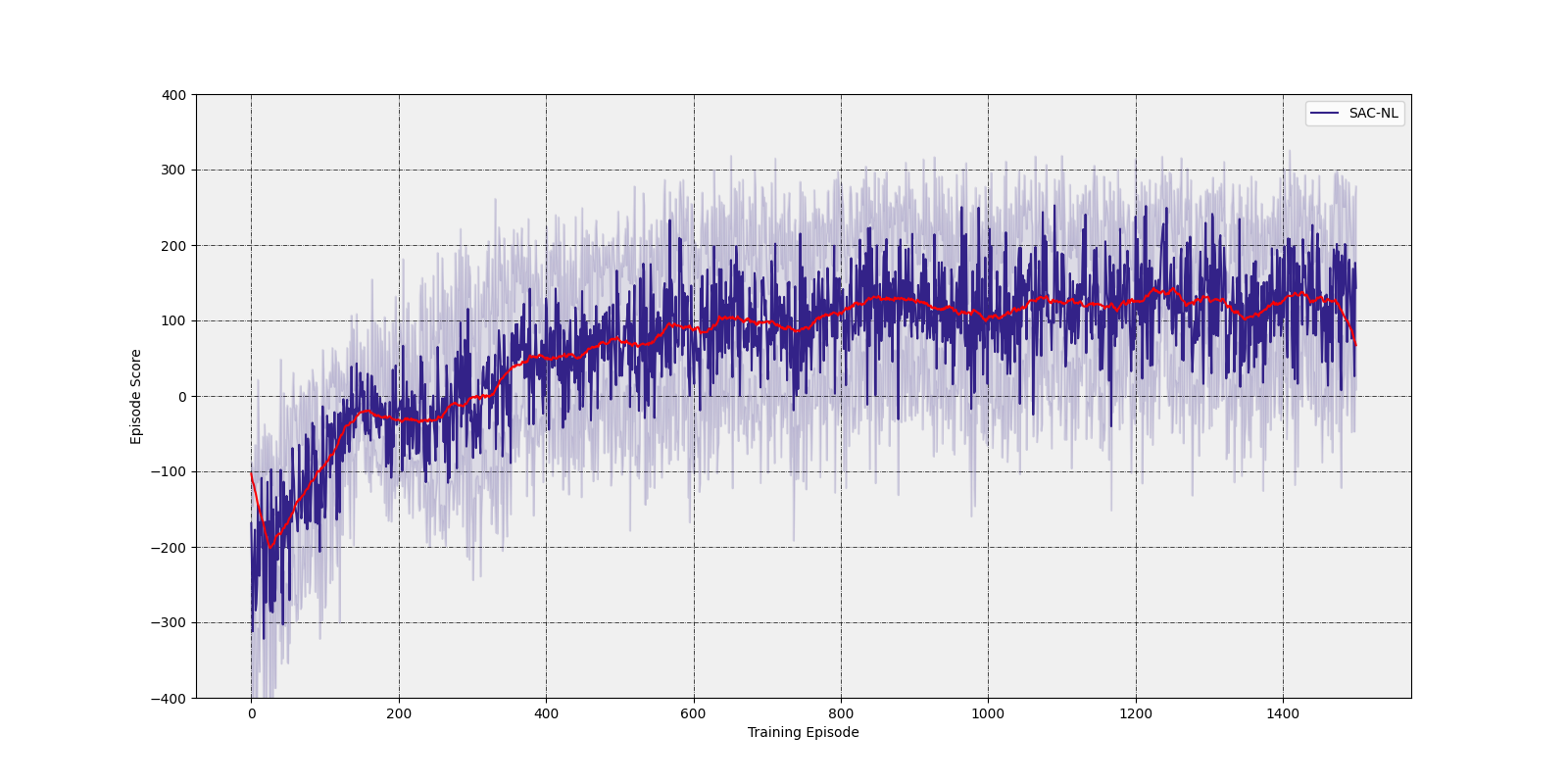}
    \caption{\scriptsize Episode scores for SAC algorithms relying on dNLRLnlc.}
    \label{fig:lunarlanderNLCseed}
\end{figure}

In Figure \ref{fig:lunarlanderNLCseed}, we can observe the results comparing the performance of the dNLRLnlc agent. The inclusion of the $sine$ is a very minor addition, but one that does take into account the state transition mechanics. Performance is better than dNLRLc in later episodes. The moving average shows mean rewards rising above $100.0$ consistently. The rules in the predicate action policy for the Lunar Lander environment are given in table \ref{Tab:LunarSAC-dNL-NLCrules}. {For each discrete action, all rules pass the parameter threshold for the disjunction layer except for a single rule associated with $fireLeft()$. For the predicate actions \textit{fireLeft()} and \textit{fireRight()}, which have four rules each, we observe the majority of definitions include the $sine$ transformed state feature $\theta$.}


\begin{table}[ht] \centering
\begin{tabular}{c l}
\hline
\scriptsize {\textbf{LunarLander}}                                                       & \scriptsize {Policy rules for dNLRLnlc}                                                                                                                               \\ \hline \small
\textbf{\scriptsize\begin{tabular}[c]{@{}c@{}}mean reward: \\ $133.4 \pm 113.9 $  \end{tabular}} & \small \begin{tabular}[c]{@{}l@{}}

\tiny \textbf{doNothing() }  \\
\tiny	 $:- (CoordX<1.66\wedge LinearVelocY<-0.61 )$\\
\tiny	 $:- ([0.90]CoordX<1.66\wedge [0.65]CoordY>-0.98\wedge [0.91]CoordY<1.42\wedge LeftLegContactFalse ))$\\
\tiny	 $:- (LinearVelocY<-0.22 )$\\
\tiny	 $:- ([0.61]CoordX<1.51\wedge AngleSine<-0.25 )$\\
\tiny	 $:- ([0.86]CoordY<2.12\wedge AngularVeloc>0.23 )$\\

\tiny \textbf{fireLeft())}  \\
\tiny	 $:- (LinearVelocX<-0.15 )$\\
\tiny	 $:- ([0.88]CoordY>-0.98\wedge LinearVelocX<0.20\wedge [0.56]LinearVelocY<0.92\wedge AngleSine>-0.08 )$\\
\tiny	 $:- ([0.53]AngleSine>-1.00\wedge AngleSine>-0.08\wedge AngularVeloc>0.07 )$\\
\tiny	 $:- [0.65] ([0.52]CoordY>-0.98\wedge CoordY<0.14\wedge [0.72]LinearVelocY>-2.32,[0.68]AngleSine>-1.00$\\
\tiny \quad $\wedge AngleSine>-0.08\wedge [0.85]AngularVeloc<1.88 )$\\
         
\tiny \textbf{fireMain() }  \\
\tiny	 $:- (LinearVelocX<-0.15\wedge LinearVelocX<-0.58\wedge [0.71]LinearVelocX<0.20\wedge AngleSine>-0.08\wedge $\\
\tiny \quad $AngularVeloc>-1.12\wedge LeftLegContactFalse )$\\
\tiny	 $:- (CoordX>-1.06\wedge LinearVelocY>-0.11 )$\\
\tiny	 $:- (CoordY>0.34\wedge LinearVelocY>-0.42\wedge [0.62]AngularVeloc>-1.12 )$\\
\tiny	 $:- (LinearVelocX>0.42\wedge AngleSine<-0.25\wedge AngularVeloc>-1.12 )$\\

\tiny\textbf{fireRight() }  \\
\tiny	 $:- (LinearVelocX>-0.25\wedge LinearVelocY<-0.61\wedge AngleSine<0.37,[0.73]AngularVeloc<2.08 )$\\
\tiny	 $:- (CoordX>0.36\wedge CoordY<0.14\wedge [0.64]LinearVelocY>-2.32\wedge [0.70]AngleSine<0.37$\\
\tiny \quad $\wedge AngleSine<0.25\wedge [0.74]AngularVeloc>-1.12 )$\\
\tiny	 $:- ([0.52]CoordY>-0.91\wedge AngleSine<0.25\wedge AngularVeloc<0.05 )$\\
\tiny	 $:- (CoordX>0.36\wedge CoordY<1.42\wedge AngleSine<0.25\wedge AngularVeloc<1.88\wedge LeftLegContactFalse )$\\

\end{tabular}                                                                                                                                                                                                                                                                                            \\ \hline
\end{tabular} 
\caption{\scriptsize The FOL policy rules for the dNLRLnlc agent, trained using SAC. Extracted continuous bounded predicates define rules for each action with the inclusion of non-linear continuous bounded predicates.}\label{Tab:LunarSAC-dNL-NLCrules}
\end{table}
\subsubsection{Discussion} 
The performance of both the dNLRLc and dNLRLnlc agents is comparable. In the CartPole problem, see Figure \ref{fig:cartpole1}, the dNLRLnlc appears more stable in later episodes than dNLRLc as evidenced by the higher occurrence of total rewards of $300$. The Lunar Lander problem, with a larger action and state space, proved more challenging for the dNLRL agents. Both agents were unable to produce mean rewards of $200$, although dNLRLnlc achieved higher rewards than dNLRLc in later episodes. While the deviation was high, indicating both dNLRLc and dNLRLnlc landed successfully on occasion, this also signified that the Lander would crash periodically. The various hyperparameters were also found to impact the model. Specifically, the instantiation of the binning scheme would sway performance. We note, that in \cite{gadgilLunarLader}, equal-width binning was not necessarily an optimal approach for the Lunar Lander except in specific regions with respect to the positional axis, and we leave this investigation for future research. Multiple trials of both agents were conducted, with performance across all episodes averaging around $100$ for dNLRLc and slightly higher for dNLRLnlc, and performance fluctuations resulted in a variety of final policy rules. The position of the lander along the $X$ and $Y$ axis, as signified by atoms associated with $CoordX$ or $CoordY$, would often have corresponding conjunction weights, indicating uncertainty in these predicates. The fluctuating bounded values during training might have caused confusion for the agent, as the position of the lander is a significant factor in landing success. To address this issue, future investigations could explore alternative schemes for training the bound weights, and consider adding additional non-linear transformation predicates or operation predicates between state features.

\section{Conclusion}


We aimed to incorporate both continuous and non-linear interpretations of dNL networks into an RRL framework, creating a dNL-ILP based agent that can learn in dynamic continuous environments. SAC was found to be the best among RL algorithms for evaluating the agent. The agent produced policies incorporating continuous and non-linear continuous predicate functions and was the first to successfully integrate ILP-based reasoning, RRL, and learning in dynamic continuous settings. The Lunar Lander problem was more challenging for the dNLRL agents, resulting in a high deviation from the mean, but our dNLRLc and dNLRLnlc agents still provide a promising starting point for ILP and RL research in continuous domains.

\bibliographystyle{eptcs}
\bibliography{generic}
\end{document}